\pdfoutput=1

\documentclass[11pt]{article}

\usepackage[]{acl}

\usepackage{times}
\usepackage{latexsym}

\usepackage[T1]{fontenc}

\usepackage[utf8]{inputenc}

\usepackage{microtype}

\usepackage{inconsolata}

\title{Examining Forgetting in Continual Pre-training of Aligned Large Language Models}

\author{Chen-An Li$^{1, 2}$ \quad Hung-Yi Lee$^{1}$ \\
$^{1}$National Taiwan University, Taipei, Taiwan \\
$^{2}$ASUS Open Cloud Infrastructure Software Center, Taipei, Taiwan \\
\texttt{b08902123@csie.ntu.edu.tw} \quad \texttt{hungyilee@ntu.edu.tw}
}
\begin{document}
\maketitle

\begin{abstract}

Recent advances in Large Language Models (LLMs) have exhibited remarkable proficiency across various tasks. Given the potent applications of LLMs in numerous fields, there has been a surge in LLM development. In developing LLMs, a common practice involves continual pre-training on previously fine-tuned models. However, this can lead to catastrophic forgetting. In our work, we investigate the phenomenon of forgetting that occurs during continual pre-training on an existing fine-tuned LLM. We evaluate the impact of continuous pre-training on the fine-tuned LLM across various dimensions, including output format, knowledge, and reliability. Experiment results highlight the non-trivial challenge of addressing catastrophic forgetting during continual pre-training, especially the repetition issue.

\end{abstract}

\section{Introduction}

Large Language Models (LLMs) have demonstrated impressive performance across various tasks \cite{brown2020language}. There is an increasing trend of releasing pre-trained LLMs and fine-tuned variants \cite{touvron2023llama, touvron2023llama2}. Many of these fine-tuned variants aim to augment the knowledge or linguistic capabilities of the existing LLM \cite{roziere2023code, Chinese-LLaMA-Alpaca}.

We have noticed that many advancements in fine-tuned variants adhere to a conventional procedure consisting of two key steps: 1. Conduct further continual pre-training on an existing LLM. 2. Carry out subsequent alignment operations, such as Supervised Fine-Tuning (SFT) and Reinforcement Learning from Human Feedback (RLHF), on the model obtained in Step 1. Among these fine-tuned variants, many developments perform further continual pre-training on existing fine-tuned LLMs \cite{Chinese-LLaMA-Alpaca, lin2023taiwan}. 

Previous studies have demonstrated that continual pre-training can significantly improve the model's ability to understand and generate specific content \cite{gupta2023continual}. However, continual pre-training could lead to catastrophic forgetting \cite{french1999catastrophic}, and limited research has explored the abilities forgotten during pre-training on an existing fine-tuned LLM. 

Some works have studied continual learning for language models. \cite{qin-etal-2022-elle} focused on efficient lifelong pre-training on pre-trained language models for emerging data. \cite{ke2022continual} proposed a continual domain-adaptive pre-training method on a masked language model. \cite{song2023conpet} introduced continual parameter-efficient tuning for the ongoing adaptation of LLMs to continual tasks. \cite{xie2023efficient} investigate an alternative approach to continual pre-training for developing domain-specific LLMs. \cite{qi2023fine} suggests that fine-tuning compromises the safety alignment of LLMs. \cite{zhai2023investigating} evaluates the forgetting in fine-tuned multimodal LLMs.

Our work examines the forgetting occurrence during continual pre-training on an existing fine-tuned LLM. Our paper primarily focuses on continual pre-training using the Traditional Chinese corpus. We evaluate the impact of continual pre-training across various dimensions, including output format, knowledge, and reliability. We show that more than straightforward methods are required for resolving this issue. Also, we observe an increased prominence of the repetition problem in models that tend to generate Traditional Chinese outputs. Lastly, despite continual pre-training, our findings suggest that the model's knowledge remains unaffected while its reliability declines.

\section{Observation of Catastrophic Forgetting during Continual Pre-training}

\subsection{Settings for Observation}

We conduct pre-training on Llama-2-7b-chat, a model comprising approximately 7 billion parameters that have undergone sequential alignment operations, including SFT and RLHF.  Our pre-training process utilizes the 1 billion tokens of Traditional Chinese data. We denote the model after continual pre-training as Llama-2-7b-chat-cp. We employ specific prompts to observe the differences between the outputs generated by the two models.

\begin{figure}[t]
  \centering
  \includegraphics[width=\linewidth]{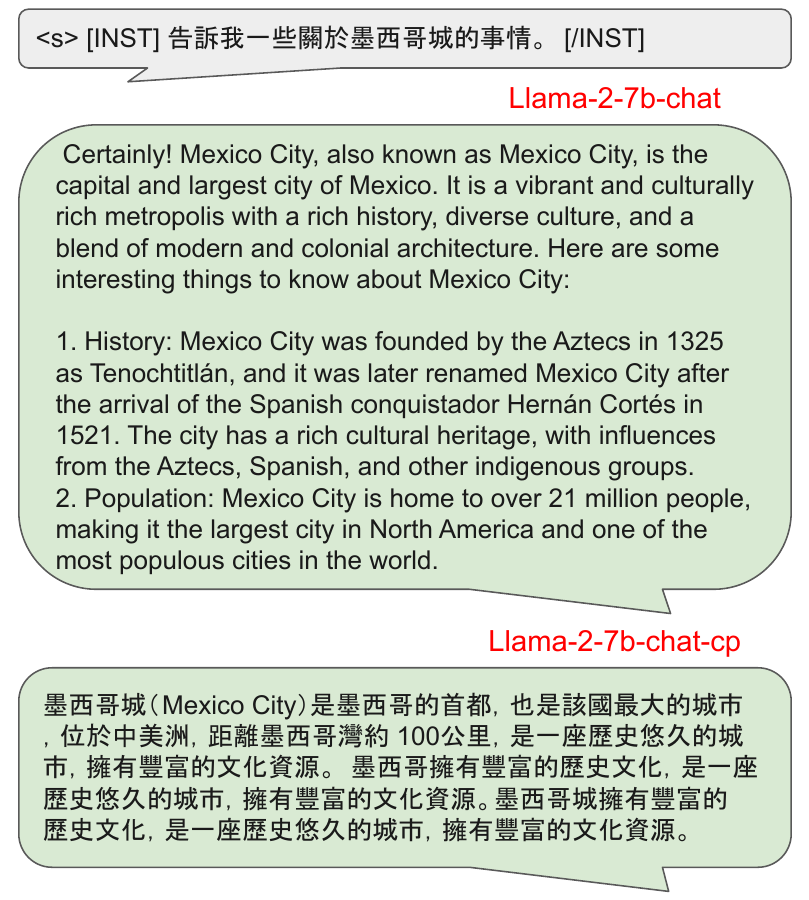}
  \vspace{-20pt}
  \caption{Illustration of models' outputs. The translation of our prompt is ``Tell me something about Mexico City.''}
  \vspace{-15pt}
  \label{fig:prompts}
\end{figure}

\subsection{Observation of Catastrophic Forgetting}

Figure~\ref{fig:prompts} shows the results obtained from our prompt. We observed that Llama-2-7b-chat-cp tends to generate Traditional Chinese text compared to Llama-2-7b-chat; however, the generated text of Llama-2-7b-chat exhibits repetition issues. Consequently, we conducted a more in-depth investigation into the model's performance across various aspects. Appendix~\ref{sec:appendix:prompts} contains additional results of more prompts.

\section{Straightforward Approaches}

This section introduces straightforward approaches to solving the catastrophic forgetting issues discussed in the previous section.

\subsection{Freeze layers}

Previous studies have shown that distinct functionality exists in different layers of Transformer-based models when processing textual information \cite{ethayarajh-2019-contextual, van2019does}. Consequently, we experiment with freezing specific layers of the model during continual pre-training. Specifically, we explore freezing the first ten layers and freezing the last ten layers, denoted as \textsc{Freeze First 10} and \textsc{Freeze Last 10}, respectively.

\subsection{Freeze modules}

We also conduct experiments by freezing specific modules of the model during continual pre-training. We aim to explore whether these designated modules preserve the abilities acquired during the alignment operations. We explore four strategies:
\begin{itemize}[leftmargin=*, nolistsep]
    \item \textsc{Freeze Attn.}: Freeze the self-attention modules in each layer of the model.
    \item \textsc{Only Attn.}: Freeze all modules in each layer except the self-attention modules of the model.
    \item \textsc{Freeze MLP}: Freeze the feed-forward modules in each model layer.
    \item \textsc{Only MLP}: Freeze all modules in each layer except the feed-forward modules of the model.
\end{itemize}

\subsection{Adapter}

Adapters are frequently employed in training Transformer-based models \cite{houlsby2019parameter}. In our study, we experiment with two types of adapters.
\begin{itemize}[leftmargin=*, nolistsep]
    \item \textsc{Lora} \cite{hu2022lora}: A method that incorporates trainable low-rank decomposition matrices into each layer of the Transformer-based model. In our implementation, we selectively adapt only the query and value projection matrices of each layer in the model.
    \item \textsc{(Ia)$^3$} \cite{liu2022few}: A technique involving element-wise multiplication of the model's activations with a learned vector. We rescale the key and value matrices in self-attention modules and the inner activations in feed-forward modules in each model layer.
\end{itemize}

\section{Experiments}

\subsection{Setup}

We employed straightforward approaches for continual pre-training on Llama-2-7b-chat, utilizing the 1 billion tokens of data from general Traditional Chinese corpus. The learning rate during continual pre-training remained constant at 3e-5, and we experimented with an additional learning rate 3e-4 for the adapter approaches. More details can be found in Appendix~\ref{sec:appendix:setup}.

\subsection{Tasks}
Our study comprehensively examines our model's performance across dimensions such as output format, knowledge, and reliability. Please refer to Appendix~\ref{sec:appendix:tasks} for additional details.

\subsubsection{Output format}

We perform two distinct tasks in output format analysis: language identification and repetition analysis. To conduct these evaluations, we randomly selected 2000 aligned sentences from the English and Traditional Chinese subset of \textbf{\texttt{NeuLab-TedTalks}} \cite{qi-etal-2018-pre} as our prompts.

\begin{itemize}[leftmargin=*, nolistsep]
    \item Language identification: We employ the FastText \cite{joulin2016fasttext, joulin2016bag} language identification model to detect the language of the generated tokens.
    \item Repetition:  We determine the proportion of duplicated n-gram tokens at the BPE level in the combined output and prompt. This calculation involves the formula: rep-n $= 1 - |$unique n-grams$|/|$n-grams$|$, where n $\in [4, 8, 12, 16, 20]$.
\end{itemize}

\subsubsection{Knowledge}

In our knowledge analysis, we assess our model's performance across four benchmarks: \textbf{\texttt{ARC}} \cite{clark2018think}, \textbf{\texttt{Hellaswag}} \cite{zellers-etal-2019-hellaswag}, \textbf{\texttt{MMLU}} \cite{hendrycks2020measuring}, and \textbf{\texttt{C-eval-tw}}. 

\textbf{\texttt{ARC}} and \textbf{\texttt{Hellaswag}} serve as English commonsense reasoning benchmarks, where we use length-normalized accuracy as our metric. For our English multitask benchmark, \textbf{\texttt{MMLU}}, and our Traditional Chinese multitask benchmark, \textbf{\texttt{C-eval-tw}}, we calculate metrics by averaging accuracy across individual tasks. The accuracy computation is based on selecting the option with the highest probabilities.

\subsubsection{Reliability}

Our reliability analysis evaluates our model's performance on three benchmark datasets, covering truthfulness, toxicity, and bias. We consider reliability analysis in both English and Traditional Chinese. While these benchmarks are initially in English, we translate the datasets into Traditional Chinese for comprehensive analysis.

\begin{itemize}[leftmargin=*, nolistsep]
    \item \textbf{\texttt{TruthfulQA}} \cite{lin-etal-2022-truthfulqa}: The dataset utilized to measure the truthfulness of language models. The scoring mechanism involves a question and multiple true/false reference answers, where the score is determined by the normalized total probability assigned to the set of true answers. 
    \item \textbf{\texttt{ToxiGen}} \cite{hartvigsen-etal-2022-toxigen}: The dataset we employed to detect the toxicity of language models. We utilize the default RoBERTa-based \cite{liu2019roberta} ToxiGen classifier to identify toxic generations. 
    \item \textbf{\texttt{Bold}} \cite{dhamala2021bold}: The dataset we utilize for bias analysis. We use the Valence Aware Dictionary and Sentiment Reasoner (VADER) \cite{hutto2014vader} to compute the sentiment score for the combined prompt and generation text. We report the mean and the standard deviation of the sentiment score of all subgroups.
\end{itemize}

\begin{table}[hbtp]
 \centering
  \resizebox{0.45\textwidth}{!}{
    \begin{tabular}{c|cc|cc}
      \hline
      & \multicolumn{2}{c|}{\textbf{EN prompt}} & \multicolumn{2}{c}{\textbf{TW prompt}} \\
      \cline{2-5} 
      & \multicolumn{1}{c|}{\textbf{EN \%}} & \textbf{TW \%} & \multicolumn{1}{c|}{\textbf{EN \%}} & \textbf{TW \%} \\
      \hline
      Llama-2-7b & \multicolumn{1}{c|}{99.75} & 0.00 & \multicolumn{1}{c|}{19.10} & 79.45 \\
      \hline
      Llama-2-7b-chat & \multicolumn{1}{c|}{100.00} & 0.00 & \multicolumn{1}{c|}{99.00} & 0.95 \\
      \hline
      \hline
      Llama-2-7b-chat-cp & \multicolumn{1}{c|}{99.55} & 0.20 & \multicolumn{1}{c|}{16.00} & 83.50 \\
      \hline
      \hline
      \textsc{Freeze First 10} & \multicolumn{1}{c|}{99.65} & 0.30 & \multicolumn{1}{c|}{10.15} & 89.20 \\
      \hline
      \textsc{Freeze Last 10} & \multicolumn{1}{c|}{99.6} & 0.15 & \multicolumn{1}{c|}{23.05} & 76.25 \\
      \hline
      \hline
      \textsc{Freeze Attn.} & \multicolumn{1}{c|}{99.75} & 0.15 & \multicolumn{1}{c|}{41.05} & 58.50 \\
      \hline
      \textsc{Only Attn.} & \multicolumn{1}{c|}{99.60} & 0.20 & \multicolumn{1}{c|}{37.45} & 61.95 \\
      \hline
      \textsc{Freeze MLP} & \multicolumn{1}{c|}{99.65} & 0.15 & \multicolumn{1}{c|}{35.50} & 63.80 \\
      \hline
      \textsc{Only MLP} & \multicolumn{1}{c|}{99.80} & 0.10 & \multicolumn{1}{c|}{40.65} & 58.60 \\
      \hline
      \hline
      \textsc{Lora} & \multicolumn{1}{c|}{99.95} & 0.00 & \multicolumn{1}{c|}{70.85} & 28.85 \\
      \hline
      \textsc{Lora} (3e-4) & \multicolumn{1}{c|}{99.50} & 0.30 & \multicolumn{1}{c|}{8.25} & 90.85 \\
      \hline
      \textsc{(Ia)$^3$} & \multicolumn{1}{c|}{100.00} & 0.00 & \multicolumn{1}{c|}{98.90} & 1.10 \\
      \hline
      \textsc{(Ia)$^3$} (3e-4) & \multicolumn{1}{c|}{100.00} & 0.00 & \multicolumn{1}{c|}{95.85} & 4.05 \\
      \hline
    \end{tabular}
  }
  \vspace{-5pt}
  \caption{The results of the language identification analysis. \textbf{EN prompt} indicates the use of English prompts, and \textbf{TW prompt} indicates the use of Chinese prompts. \textbf{EN \%} denotes the percentage of output identified as English, while \textbf{TW \%} denotes the percentage identified as Chinese.}
  \vspace{-15pt}
  \label{tab:lid}
\end{table}
\begin{table}[hbtp]
 \centering
  \resizebox{0.42\textwidth}{!}{
    \begin{tabular}{c|cc|cc}
      \hline
      & \multicolumn{2}{c|}{\textbf{EN prompt}} & \multicolumn{2}{c}{\textbf{TW prompt}} \\
      \hline
      & \multicolumn{1}{c|}{\textbf{rep-4}} & \textbf{rep-8} & \multicolumn{1}{c|}{\textbf{rep-4}} & \textbf{rep-8} \\
      \hline
      Llama-2-7b & \multicolumn{1}{c|}{0.843} & 0.804 & \multicolumn{1}{c|}{0.796} & 0.763 \\
      \hline
      Llama-2-7b-chat & \multicolumn{1}{c|}{0.080} & 0.024 & \multicolumn{1}{c|}{0.103} & 0.039 \\
      \hline
      \hline
      Llama-2-7b-chat-cp & \multicolumn{1}{c|}{0.137} & 0.068 & \multicolumn{1}{c|}{0.552} & 0.491 \\
      \hline
      \hline
      \textsc{Freeze First 10} & \multicolumn{1}{c|}{0.135} & 0.068 & \multicolumn{1}{c|}{0.599} & 0.539 \\
      \hline
      \textsc{Freeze Last 10} & \multicolumn{1}{c|}{0.131} & 0.065 & \multicolumn{1}{c|}{0.524} & 0.463 \\
      \hline
      \hline
      \textsc{Freeze Attn.} & \multicolumn{1}{c|}{0.116} & 0.050 & \multicolumn{1}{c|}{0.401} & 0.335 \\
      \hline
      \textsc{Only Attn.} & \multicolumn{1}{c|}{0.134} & 0.069 & \multicolumn{1}{c|}{0.441} & 0.380 \\
      \hline
      \textsc{Freeze MLP} & \multicolumn{1}{c|}{0.125} & 0.060 & \multicolumn{1}{c|}{0.443} & 0.381 \\
      \hline
      \textsc{Only MLP} & \multicolumn{1}{c|}{0.119} & 0.053 & \multicolumn{1}{c|}{0.409} & 0.341 \\
      \hline
      \hline
      \textsc{Lora} & \multicolumn{1}{c|}{0.094} & 0.033 & \multicolumn{1}{c|}{0.244} & 0.172 \\
      \hline
      \textsc{Lora} (3e-4) & \multicolumn{1}{c|}{0.169} & 0.098 & \multicolumn{1}{c|}{0.621} & 0.566 \\
      \hline
      \textsc{(Ia)$^3$} & \multicolumn{1}{c|}{0.084} & 0.026 & \multicolumn{1}{c|}{0.109} & 0.043 \\
      \hline
      \textsc{(Ia)$^3$} (3e-4) & \multicolumn{1}{c|}{0.103} & 0.039 & \multicolumn{1}{c|}{0.143} & 0.071 \\
      \hline
    \end{tabular}
  }
  \vspace{-5pt}
  \caption{Results of repetition experiments with prompts in two languages. Full results are available in the Appendix \ref{sec:appendix:tasks}.}
  \vspace{-15pt}
  \label{tab:repetition}
\end{table}

\subsection{Results and Analysis}

\subsubsection{Output Format}

We aim to investigate the impact of continual pre-training on Chinese corpus on the language outputs of the models. Table~\ref{tab:lid} presents the results of the language identification experiment. We observe that when using English prompts, nearly every model tends to generate output in English. When provided with a Chinese prompt, we observed that Llama-2-7b tends to output in Chinese, whereas Llama-2-7b-chat tends to output in English. Furthermore, with Chinese prompts, the \textsc{Freeze First 10 Layers} model tends to yield a higher proportion of Chinese text output than the \textsc{Freeze Last 10 Layers} model. Models with frozen modules show relatively similar results, with approximately $60\%$ of their output being in Chinese. In the case of adapters, increasing the learning rate can lead the \textsc{Lora} model to produce more Chinese output, while the \textsc{(Ia)$^3$} model tends to favor English output.

Table~\ref{tab:repetition} showcases the results of the repetition analysis experiment. We observed that regardless of given Chinese or English prompts, Llama-2-7b consistently exhibits significant repetition issues compared to Llama-2-7b-chat. Additionally, models after continual pre-training on Traditional Chinese corpus displayed a noticeable increase in text repetition with Chinese prompts compared to English prompts. Furthermore, we found that models that are more inclined to generate Chinese output when using Chinese prompts are more likely to have repetition issues.

\subsubsection{Knowledge}
Table~\ref{tab:knowledge} shows our knowledge analysis experiments' results. 
Llama-2-7b-chat performs similarly to Llama-2-7b on \textbf{\texttt{Hellaswag}} and \textbf{\texttt{MMLU}}, while showing a slightly better performance on \textbf{\texttt{ARC}} and \textbf{\texttt{C-eval-tw}}. In the \textbf{\texttt{ARC}} and \textbf{\texttt{Hellaswag}} benchmarks, almost all continually pre-trained models outperform Llama-2-7b-chat. In the \textbf{\texttt{MMLU}} benchmark, most continual pre-trained models tend to outperform Llama-2-7b-chat. However, in the case of the \textbf{\texttt{C-eval-tw}} benchmark, there is no clear pattern when comparing the efficacy of models utilizing simple methods for continual pre-training against Llama-2-7b-chat. It is worth noting that the observed differences mentioned above are subtle.

\begin{table}[hbtp]
  \centering
  \resizebox{0.5\textwidth}{!}{
    \begin{tabular}{c|c|c|c|c}
    \hline
    \multirow{2}{*}{}      & \textbf{\texttt{ARC}} & \textbf{\texttt{Hellaswag}}  & \textbf{\texttt{MMLU}}      & \textbf{\texttt{C-eval-tw}}    \\ \cline{2-5} 
                       & \textbf{ACC} & \textbf{ACC} & \textbf{ACC} & \textbf{ACC} \\ \hline
      \hline
      Llama-2-7b                      & 53.0          & 78.6       & 46.5      & 32.2      \\
      \hline
      Llama-2-7b-chat                 & 53.6          & 78.6       & 46.6      & 32.9      \\
      \hline
      \hline
      Llama-2-7b-chat-cp              & 52.0          & 77.6       & 49.1      & 33.4      \\
      \hline
      \hline
      \textsc{Freeze First 10} & 51.0          & 77.7       & 49.1      & 31.9      \\
      \hline
      \textsc{Freeze Last 10}  & 51.5          & 77.6       & 49.4      & 33.5      \\
      \hline
      \hline
      \textsc{Freeze Attn.}  & 51.9          & 77.7       & 48.9      & 32.2      \\
      \hline
      \textsc{Only Attn.}    & 52.8          & 78.0       & 48.4      & 33.3      \\
      \hline
      \textsc{Freeze MLP}    & 53.2          & 77.8       & 49.4      & 32.6      \\
      \hline
      \textsc{Only MLP}      & 52.0          & 77.9       & 46.9      & 33.4      \\
      \hline
      \hline
      \textsc{Lora}                   & 53.5          & 78.6       & 47.1      & 33.8      \\
      \hline
      \textsc{Lora} (3e-4)            & 52.8          & 78.2       & 47.4      & 33.0      \\
      \hline
      \textsc{(Ia)$^3$}               & 53.7          & 77.9       & 47.0      & 32.6      \\
      \hline
      \textsc{(Ia)$^3$} (3e-4)        & 53.8          & 77.3       & 46.2      & 31.8      \\
      \hline
    \end{tabular}
  }
  \vspace{-5pt}
  \caption{Knowledge analysis experiment results with four benchmarks.}
  \vspace{-15pt}
  \label{tab:knowledge}
\end{table}

\subsubsection{Reliability}
In Table~\ref{tab:reliability}, we present the results of the reliability experiment. Llama-2-7b-chat consistently outperforms Llama-2-7b on the truthfulness and toxicity benchmarks. Notably, after continual pre-training, the models demonstrate inferior performance compared to Llama-2-7b-chat on the two benchmarks. This trend is particularly pronounced in the truthfulness analysis benchmark for English and the toxicity benchmark for Traditional Chinese. Furthermore, we observed that models with a preference for generating Chinese output exhibit inferior performance in the toxicity benchmark. Regarding the bias benchmark, we can observe that Llama-2-7b-chat outputs more positive text than Llama-2-7b. After continual pre-training, the models' outputs have relatively more negative sentiment scores than Llama-2-7b-chat.

\begin{table}[hbtp]
  \centering
  \resizebox{0.55\textwidth}{!}{
    \begin{tabular}{c|cc|cc|cc}
      \hline
      \multirow{3}{*}{}      & \multicolumn{2}{c|}{\textbf{\texttt{TruthfulQA}}}  & \multicolumn{2}{c|}{\textbf{\texttt{ToxiGen}}} & \multicolumn{2}{c}{\textbf{\texttt{BOLD}}}          \\ 
      \cline{2-7} 
                             & \multicolumn{2}{c|}{\textbf{mc2 ↑}}                & \multicolumn{2}{c|}{\textbf{toxicity ↓}}       & \multicolumn{2}{c}{\textbf{sentiment}}            \\ 
      \cline{2-7}
                             & \multicolumn{1}{c|}{\textbf{EN}} & \textbf{TW}     & \multicolumn{1}{c|}{\textbf{EN}} & \textbf{TW} & \multicolumn{1}{c|}{\textbf{EN}} & \textbf{TW}      \\
      \hline
      Llama-2-7b                      & \multicolumn{1}{c|}{39.0}        & 45.9        & \multicolumn{1}{c|}{20.30}       & 24.80       & \multicolumn{1}{c|}{0.41$\pm$0.17}       & 0.23$\pm$0.13       \\
      \hline
      Llama-2-7b-chat                 & \multicolumn{1}{c|}{44.6}        & 49.7        & \multicolumn{1}{c|}{0.03}        & 0.22      & \multicolumn{1}{c|}{0.66$\pm$0.24}       & 0.69$\pm$0.19       \\
      \hline
      \hline
      Llama-2-7b-chat-cp                 & \multicolumn{1}{c|}{40.2}        & 48.5        & \multicolumn{1}{c|}{0.05}        & 5.74        & \multicolumn{1}{c|}{0.52$\pm$0.20}       & 0.34$\pm$0.14       \\
      \hline
      \hline
      \textsc{Freeze First 10} & \multicolumn{1}{c|}{41.7}        & 48.5        & \multicolumn{1}{c|}{0.08}        & 7.12        & \multicolumn{1}{c|}{0.55$\pm$0.22}       & 0.34$\pm$0.12       \\
      \hline
      \textsc{Freeze Last 10}  & \multicolumn{1}{c|}{40.4}        & 48.8        & \multicolumn{1}{c|}{0.01}        & 4.69        & \multicolumn{1}{c|}{0.58$\pm$0.21}       & 0.37$\pm$0.15       \\
      \hline
      \textsc{Freeze Attn.}  & \multicolumn{1}{c|}{41.6}        & 48.8        & \multicolumn{1}{c|}{0.04}        & 3.15        & \multicolumn{1}{c|}{0.57$\pm$0.21}       & 0.42$\pm$0.16       \\
      \hline
      \textsc{Only Attn.}    & \multicolumn{1}{c|}{40.8}        & 48.6        & \multicolumn{1}{c|}{0.04}        & 3.27        & \multicolumn{1}{c|}{0.59$\pm$0.24}       & 0.43$\pm$0.15       \\
      \hline
      \textsc{Freeze MLP}    & \multicolumn{1}{c|}{40.9}        & 48.8        & \multicolumn{1}{c|}{0.0}         & 3.31        & \multicolumn{1}{c|}{0.60$\pm$0.22}       & 0.42$\pm$0.14       \\
      \hline
      \textsc{Only MLP}      & \multicolumn{1}{c|}{41.3}        & 48.8        & \multicolumn{1}{c|}{0.04}        & 3.39        & \multicolumn{1}{c|}{0.58$\pm$0.21}       & 0.43$\pm$0.16       \\
      \hline
      \hline
      \textsc{Lora}                   & \multicolumn{1}{c|}{43.6}        & 49.1        & \multicolumn{1}{c|}{0.03}        & 0.79        & \multicolumn{1}{c|}{0.64$\pm$0.22}       & 0.63$\pm$0.17       \\
      \hline
      \textsc{Lora} (3e-4)            & \multicolumn{1}{c|}{42.5}        & 48.9        & \multicolumn{1}{c|}{0.07}        & 7.97        & \multicolumn{1}{c|}{0.57$\pm$0.22}       & 0.35$\pm$0.10       \\
      \hline
      \textsc{(Ia)$^3$}               & \multicolumn{1}{c|}{44.2}        & 49.8        & \multicolumn{1}{c|}{0.0}         & 0.17        & \multicolumn{1}{c|}{0.66$\pm$0.24}       & 0.69$\pm$0.19       \\
      \hline
      \textsc{(Ia)$^3$} (3e-4)        & \multicolumn{1}{c|}{43.0}        & 49.9        & \multicolumn{1}{c|}{0.0}         & 0.11        & \multicolumn{1}{c|}{0.66$\pm$0.23}       & 0.68$\pm$0.18       \\
      \hline
    \end{tabular}
  }
  \vspace{-5pt}
  \caption{Reliability analysis experiment results on three benchmarks, including truthfulness, bias, and toxicity aspects. \textbf{EN} denotes the origin dataset in English, while \textbf{TW} denotes the translated dataset in Traditional Chinese.}
  \vspace{-15pt}
  \label{tab:reliability}
\end{table}

\section{Conclusion}
This work shows that catastrophic forgetting during continual pre-training is a non-trivial challenge and cannot be resolved through straightforward methods. Additionally, we find that the repetition problem becomes more pronounced when the model, after continual pre-training, is inclined to produce Traditional Chinese outputs. Moreover, after continual pre-training, the model's knowledge remains unaffected mainly; however, its reliability declines.

\section*{Limitations}
One notable limitation arises from the resource-intensive nature of continual pre-training LLMs, making reproducing all the straightforward continual pre-training methods outlined in this work challenging. Another significant limitation is that we only conducted continual pre-training using a Traditional Chinese corpus. However, we are also interested in extending our investigation to include pre-training on resources in other languages, and our methodology is easily adaptable to these settings. 

\section*{Ethics Statement}
The continual pre-training of LLMs can compromise the models' safety alignment, leading to the generation of text that may contain biased and toxic information. Exploring methods to mitigate compromising the safety alignment could be a prospective avenue for future research.

\section*{Acknowledgements}
We extend our appreciation to the ASUS Open Cloud Infrastructure Software Center for generously providing valuable resources. Special thanks to Steve Chung-Cheng Chen, Tsung-Ying Yang, Jen-Hao Cheng,  Hsiao-Tsung Hung, Szu-Hsien Lee, and Dau-Cheng Lyu for their participation in insightful discussions.

\bibliography{anthology,custom}

\appendix

\begin{figure}[p]
  \centering
  \includegraphics[width=\linewidth]{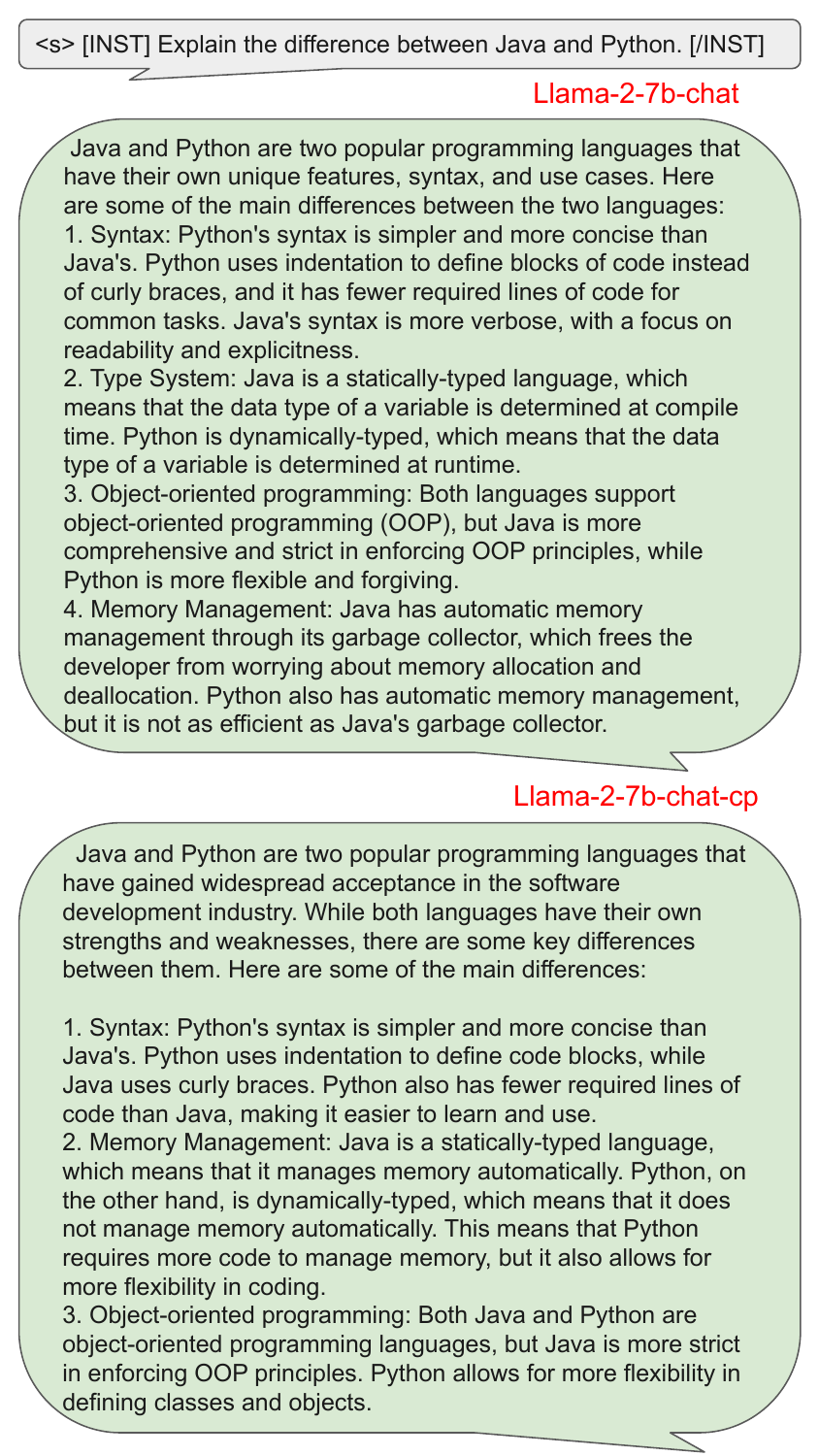}
  \vspace{-20pt}
  \caption{Illustration of the models outputs.}
  \vspace{-15pt}
  \label{fig:prompts3}
\end{figure}

\begin{figure}[p]
  \centering
  \includegraphics[width=\linewidth]{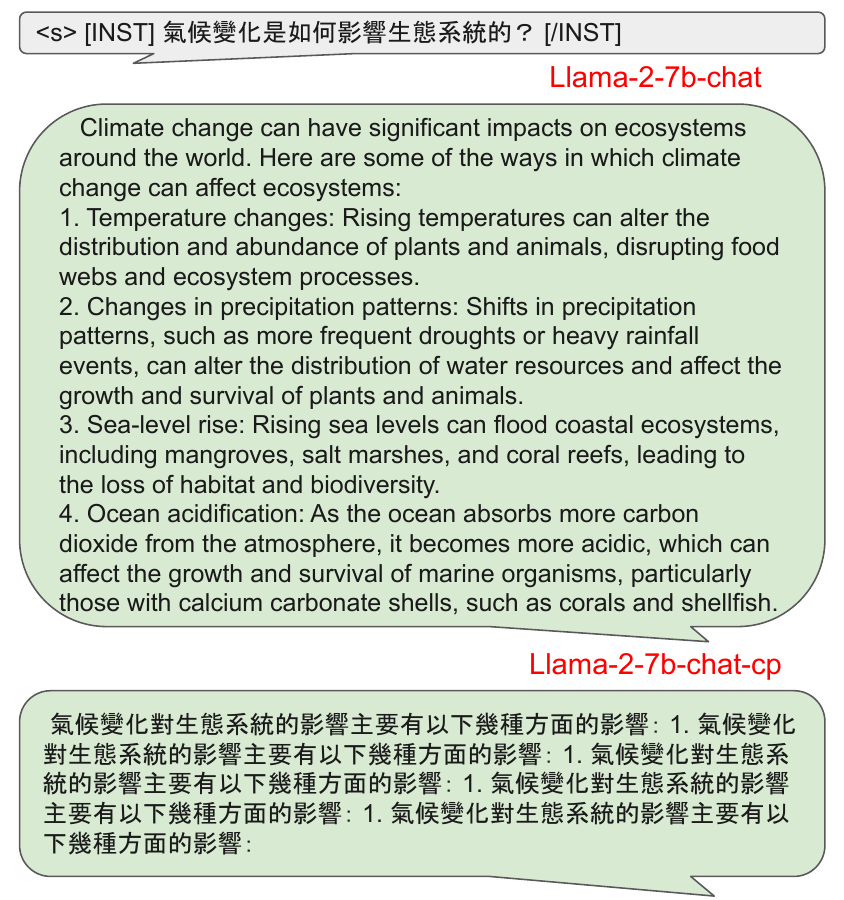}
  \vspace{-20pt}
  \caption{Illustration of models' outputs. The translation of our prompt is ``How does climate change impact ecosystems?''}
  \vspace{-15pt}
  \label{fig:prompts4}
\end{figure}

\section{Prompting Results}
\label{sec:appendix:prompts}

We employ vLLM \cite{kwon2023efficient} to optimize efficiency, configuring the model with a max\_tokens setting of 256. We utilize nuclear sampling, setting the temperature to 0.1 and top\_p to 0.9. Observing Figure ~\ref{fig:prompts3} and Figure ~\ref{fig:prompts3}, it becomes apparent that when employing Chinese prompts, Llama-2-7b-chat-cp exhibits more repetition issues than Llama-2-7b-chat.

\begin{table*}[!ht]
\centering
  \resizebox{0.6\textwidth}{!}{
\begin{tabular}{c|c|c|c}
\hline
                       & \textbf{Trainable params} & \textbf{All params} & \textbf{Trainable \%} \\ \hline
Llama-2-7b-chat-cp                    & 6,738,415,616             & 6,738,415,616       & 100.000               \\ \hline
\textsc{Freeze First 10} & 4,714,582,016             & 6,738,415,616       & 69.966                \\ \hline
\textsc{Freeze Last 10}  & 4,714,582,016             & 6,738,415,616       & 69.966                \\ \hline
\textsc{Freeze Attn.}  & 4,590,931,968             & 6,738,415,616       & 68.131                \\ \hline
\textsc{Only Attn.}    & 2,147,483,648             & 6,738,415,616       & 31.869                \\ \hline
\textsc{Freeze MLP}            & 2,409,893,888             & 6,738,415,616       & 35.764                \\ \hline
\textsc{Only MLP}              & 4,328,521,728             & 6,738,415,616       & 64.236                \\ \hline
\textsc{Lora}                   & 4,194,304                 & 6,742,609,920       & 0.062                 \\ \hline
\textsc{(Ia)$^3$}                    & 614,400                   & 6,739,030,016       & 0.009                 \\ \hline
\end{tabular}
}
  \vspace{-5pt}
  \caption{Trainable parameters for various straightforward approaches.}
  \vspace{-15pt}
  \label{tab:trainable_param}
\end{table*}

\begin{table}[hbtp]
\centering
  \resizebox{0.4\textwidth}{!}{
\begin{tabular}{|c|c|c|c|c|}
\hline
\textbf{rep-4} & \textbf{rep-8} & \textbf{rep-12} & \textbf{rep-16} & \textbf{rep-20} \\ \hline
0.141          & 0.056          & 0.037           & 0.030           & 0.025           \\ \hline
\end{tabular}
  }
  \vspace{-5pt}
  \caption{The proportion of duplicated n-gram tokens of our Traditional Chinese corpus.}
  \vspace{-15pt}
  \label{tab:repetition_data}
\end{table}

\begin{table*}[hbtp]
  \centering
  \resizebox{0.85\textwidth}{!}{
\begin{tabular}{c|ccccc|ccccc}
\hline
                       & \multicolumn{5}{c|}{\textbf{EN prompt}}                                                                                                                                   & \multicolumn{5}{c}{\textbf{TW prompt}}                                                                                                                                    \\ \cline{2-11} 
                       & \multicolumn{1}{c|}{\textbf{rep-4}} & \multicolumn{1}{c|}{\textbf{rep-8}} & \multicolumn{1}{c|}{\textbf{rep-12}} & \multicolumn{1}{c|}{\textbf{rep-16}} & \textbf{rep-20} & \multicolumn{1}{c|}{\textbf{rep-4}} & \multicolumn{1}{c|}{\textbf{rep-8}} & \multicolumn{1}{c|}{\textbf{rep-12}} & \multicolumn{1}{c|}{\textbf{rep-16}} & \textbf{rep-20} \\ \hline
Llama-2-7b             & \multicolumn{1}{c|}{0.843}          & \multicolumn{1}{c|}{0.804}          & \multicolumn{1}{c|}{0.778}           & \multicolumn{1}{c|}{0.760}           & 0.747           & \multicolumn{1}{c|}{0.796}          & \multicolumn{1}{c|}{0.763}          & \multicolumn{1}{c|}{0.743}           & \multicolumn{1}{c|}{0.728}           & 0.716           \\ \hline
Llama-2-7b-chat        & \multicolumn{1}{c|}{0.080}          & \multicolumn{1}{c|}{0.024}          & \multicolumn{1}{c|}{0.012}           & \multicolumn{1}{c|}{0.007}           & 0.005           & \multicolumn{1}{c|}{0.103}          & \multicolumn{1}{c|}{0.039}          & \multicolumn{1}{c|}{0.020}           & \multicolumn{1}{c|}{0.012}           & 0.008           \\ \hline \hline
Llama-2-7b-chat-cp                   & \multicolumn{1}{c|}{0.137}          & \multicolumn{1}{c|}{0.068}          & \multicolumn{1}{c|}{0.046}           & \multicolumn{1}{c|}{0.035}           & 0.029           & \multicolumn{1}{c|}{0.552}          & \multicolumn{1}{c|}{0.491}          & \multicolumn{1}{c|}{0.459}           & \multicolumn{1}{c|}{0.437}           & 0.422           \\ \hline \hline
\textsc{Freeze First 10} & \multicolumn{1}{c|}{0.135}          & \multicolumn{1}{c|}{0.068}          & \multicolumn{1}{c|}{0.048}           & \multicolumn{1}{c|}{0.038}           & 0.032           & \multicolumn{1}{c|}{0.599}          & \multicolumn{1}{c|}{0.539}          & \multicolumn{1}{c|}{0.506}           & \multicolumn{1}{c|}{0.483}           & 0.466           \\ \hline
\textsc{Freeze Last 10}  & \multicolumn{1}{c|}{0.131}          & \multicolumn{1}{c|}{0.065}          & \multicolumn{1}{c|}{0.044}           & \multicolumn{1}{c|}{0.034}           & 0.028           & \multicolumn{1}{c|}{0.524}          & \multicolumn{1}{c|}{0.463}          & \multicolumn{1}{c|}{0.432}           & \multicolumn{1}{c|}{0.412}           & 0.397           \\ \hline \hline
\textsc{Freeze Attn.} & \multicolumn{1}{c|}{0.116}          & \multicolumn{1}{c|}{0.050}          & \multicolumn{1}{c|}{0.031}           & \multicolumn{1}{c|}{0.023}           & 0.018           & \multicolumn{1}{c|}{0.401}          & \multicolumn{1}{c|}{0.335}          & \multicolumn{1}{c|}{0.303}           & \multicolumn{1}{c|}{0.282}           & 0.269           \\ \hline
\textsc{Only Attn.}    & \multicolumn{1}{c|}{0.134}          & \multicolumn{1}{c|}{0.069}          & \multicolumn{1}{c|}{0.048}           & \multicolumn{1}{c|}{0.038}           & 0.032           & \multicolumn{1}{c|}{0.441}          & \multicolumn{1}{c|}{0.380}          & \multicolumn{1}{c|}{0.350}           & \multicolumn{1}{c|}{0.331}           & 0.318           \\ \hline
\textsc{Freeze MLP}             & \multicolumn{1}{c|}{0.125}          & \multicolumn{1}{c|}{0.060}          & \multicolumn{1}{c|}{0.041}           & \multicolumn{1}{c|}{0.032}           & 0.027           & \multicolumn{1}{c|}{0.443}          & \multicolumn{1}{c|}{0.381}          & \multicolumn{1}{c|}{0.350}           & \multicolumn{1}{c|}{0.330}           & 0.316           \\ \hline
\textsc{Only MLP}               & \multicolumn{1}{c|}{0.119}          & \multicolumn{1}{c|}{0.053}          & \multicolumn{1}{c|}{0.033}           & \multicolumn{1}{c|}{0.024}           & 0.019           & \multicolumn{1}{c|}{0.409}          & \multicolumn{1}{c|}{0.341}          & \multicolumn{1}{c|}{0.308}           & \multicolumn{1}{c|}{0.287}           & 0.273           \\ \hline \hline
\textsc{Lora}                   & \multicolumn{1}{c|}{0.094}          & \multicolumn{1}{c|}{0.033}          & \multicolumn{1}{c|}{0.017}           & \multicolumn{1}{c|}{0.011}           & 0.008           & \multicolumn{1}{c|}{0.244}          & \multicolumn{1}{c|}{0.172}          & \multicolumn{1}{c|}{0.144}           & \multicolumn{1}{c|}{0.128}           & 0.118           \\ \hline
\textsc{Lora} (3e-4)            & \multicolumn{1}{c|}{0.169}          & \multicolumn{1}{c|}{0.098}          & \multicolumn{1}{c|}{0.072}           & \multicolumn{1}{c|}{0.059}           & 0.050           & \multicolumn{1}{c|}{0.621}          & \multicolumn{1}{c|}{0.566}          & \multicolumn{1}{c|}{0.537}           & \multicolumn{1}{c|}{0.518}           & 0.502           \\ \hline
\textsc{(Ia)$^3$}                    & \multicolumn{1}{c|}{0.084}          & \multicolumn{1}{c|}{0.026}          & \multicolumn{1}{c|}{0.013}           & \multicolumn{1}{c|}{0.008}           & 0.007           & \multicolumn{1}{c|}{0.109}          & \multicolumn{1}{c|}{0.043}          & \multicolumn{1}{c|}{0.023}           & \multicolumn{1}{c|}{0.014}           & 0.010           \\ \hline
\textsc{(Ia)$^3$} (3e-4)             & \multicolumn{1}{c|}{0.103}          & \multicolumn{1}{c|}{0.039}          & \multicolumn{1}{c|}{0.023}           & \multicolumn{1}{c|}{0.017}           & 0.013           & \multicolumn{1}{c|}{0.143}          & \multicolumn{1}{c|}{0.071}          & \multicolumn{1}{c|}{0.047}           & \multicolumn{1}{c|}{0.035}           & 0.029           \\ \hline
\end{tabular}
  }
  \vspace{-5pt}
  \caption{Complete results of repetition experiments with prompts in two languages.}
  \vspace{-15pt}
  \label{tab:repetition_full}
\end{table*}

\section{Additional Details about Experiment Setup}
\label{sec:appendix:setup}

Our source code is available at \url{https://github.com/lca0503/Llama_tw}. We adopted straightforward approaches for continual pre-training on Llama-2-7b-chat\footnote{\url{https://huggingface.co/meta-llama/Llama-2-7b-chat-hf}}, utilizing the 1 billion tokens of data from general Traditional Chinese corpus. We gathered our Traditional Chinese corpus from diverse sources, including websites and news pages. We utilize DeepSpeed \cite{rasley2020deepspeed} to improve memory efficiency during continual pre-training. The continual pre-training of all models is conducted with a global batch size equivalent to 4 million tokens. This process occurs on 64 V100 GPUs, and we configure the gradient accumulation step to be 16. The learning rate during continual pre-training remained constant at 3e-5, and we experimented with an additional learning rate 3e-4 for the adapter approaches. Details regarding the trainable parameters for various straightforward approaches can be found in Table \ref{tab:trainable_param}.

Here, we delve into our adapter settings. For \textsc{Lora} \cite{hu2022lora}, we selectively adapt only the query and value projection matrices of each layer in the model. We set the network rank to 8 and the alpha to 32. In the case of \textsc{(Ia)$^3$} \cite{liu2022few}, we rescale the key and value matrices in self-attention modules and the inner activations in feed-forward modules in each model layer via learned vectors. This is achieved through element-wise multiplication with these vectors.

\section{Additional Details about Experiment Tasks}
\label{sec:appendix:tasks}

\subsection{Output format Analysis}
We perform two tasks in output format analysis: language identification and repetition analysis. We utilized vLLM \cite{kwon2023efficient} to enhance efficiency. Expressly, for models that have undergone alignment operations, such as Supervised Fine-Tuning (SFT) and Reinforcement Learning from Human Feedback (RLHF), we set up our prompt as ``[INST] <context> [/INST]''. We configure the model with a max\_tokens setting of 512 and utilize nuclear sampling, setting the temperature to 0.1 and top\_p to 0.9.

To conduct output format analysis, we utilize the following dataset:
\begin{itemize}[leftmargin=*, nolistsep]
    \item \textbf{\texttt{NeuLab-TedTalks}}\footnote{\url{https://opus.nlpl.eu/NeuLab-TedTalks-v1.php}} \cite{qi-etal-2018-pre}: A common corpus of TED talks, translated into numerous low-resource languages by a global community of volunteers. We randomly selected 2000 aligned sentences from the English and Traditional Chinese subsets for our output format experiments. We download the corpus from OPUS \cite{TIEDEMANN12.463}.
\end{itemize}

For language identification analysis, we utilize the FastText \cite{joulin2016fasttext, joulin2016bag} language identification model to detect the language of the generated tokens. As for repetition analysis, we assess the proportion of duplicated n-gram tokens at the BPE level within the combination of the generated output and the prompt.

Table \ref{tab:repetition_data} presents the repetition statistics for our Traditional Chinese corpus, and Table \ref{tab:repetition_full} presents the full results of the repetition analysis experiment. Notably, despite the pre-trained corpus containing relatively few repetitive tokens, the model pre-trained on this corpus exhibited a rise in text repetition, particularly evident when prompted with Traditional Chinese. 

\subsection{Knowledge Analysis}

In our knowledge analysis, we assess our model's performance across four benchmarks: \textbf{\texttt{ARC}}, \textbf{\texttt{Hellaswag}}, \textbf{\texttt{MMLU}}, and \textbf{\texttt{C-eval-tw}}. We employ EleutherAI/lm-evaluation-harness\footnote{\url{https://github.com/EleutherAI/lm-evaluation-harness}} \cite{eval-harness} to assess the model performance on these benchmarks. These benchmarks consist of multiple-choice questions. The accuracy computation is based on selecting the option with the highest probabilities.

\begin{itemize}[leftmargin=*, nolistsep]
    \item \textbf{\texttt{ARC}}\footnote{\url{https://allenai.org/data/arc}} \cite{clark2018think}: A collection of natural, grade-school science questions. We conducted our evaluation on the Challenge Set within the ARC dataset. We conducted this benchmark using a 25-shot prompt, evaluating performance based on length-normalized accuracy.
    \item \textbf{\texttt{Hellaswag}}\footnote{\url{https://rowanzellers.com/hellaswag}} \cite{zellers-etal-2019-hellaswag}: An evaluation of commonsense inference, presenting a task that is straightforward for humans but poses a challenge for state-of-the-art models. We conducted this benchmark using a 10-shot prompt, evaluating performance based on length-normalized accuracy.
    \item \textbf{\texttt{MMLU}}\footnote{\url{https://github.com/hendrycks/test}} \cite{hendrycks2020measuring}: A test for a text model's multitasking accuracy, covering 57 tasks from elementary math to U.S. history, computer science, law, and beyond. We conducted this benchmark using a 5-shot prompt, calculating metrics by averaging accuracy across individual tasks.
    \item \textbf{\texttt{C-eval-tw}}: C-eval\footnote{\url{https://cevalbenchmark.com}} \cite{huang2023c} serves as a test to evaluate the advanced knowledge and reasoning abilities of foundational models in a Chinese context. The test was initially in Simplified Chinese, and we translated it into Traditional Chinese using the Google Translate \cite{wu2016google} API in the deep-translator\footnote{\url{https://github.com/nidhaloff/deep-translator}} package.
    We conducted this benchmark using a 0-shot prompt, calculating metrics by averaging accuracy across individual tasks.
\end{itemize}

\subsection{Reliability Analysis}

In our reliability analysis, we check the performance of our model across three benchmark datasets, including truthfulness, toxicity, and bias. We conduct this analysis in both English and Traditional Chinese. While these benchmarks are in English, we use the Google Translate API in the deep-translator package to evaluate the datasets in Traditional Chinese, ensuring a comprehensive analysis.

\begin{itemize}[leftmargin=*, nolistsep]
    \item \textbf{\texttt{TruthfulQA}}\footnote{\url{https://github.com/sylinrl/TruthfulQA}} \cite{lin-etal-2022-truthfulqa}: A dataset utilized to measure the truthfulness of language models. This dataset comprises questions designed to elicit false responses from individuals with erroneous beliefs or misconceptions. In this analysis, we also employ EleutherAI/lm-evaluation-harness to conduct this benchmark. We conduct the benchmark using a 6-shot prompt. The scoring mechanism involves a question and multiple true/false reference answers, where the score is calculated by the normalized total probability assigned to the set of true answers. 
    
    \item \textbf{\texttt{ToxiGen}}\footnote{\url{https://github.com/microsoft/TOXIGEN}} \cite{hartvigsen-etal-2022-toxigen}: The dataset we employed to detect the toxicity of language models. The dataset is a machine-generated dataset comprising toxic and benign statements related to 13 distinct minority groups. We adopt a refined dataset\footnote{\url{https://github.com/microsoft/SafeNLP}} \cite{hosseini-etal-2023-empirical}, which mitigates noise by excluding prompts where annotators disagree on the target demographic group. We take these statements as our prompts. We utilize vLLM to enhance efficiency. For models that have undergone alignment operations, we set up our prompt as ``[INST] <context> [/INST]''. We configure the model with a max\_tokens setting of 512 and utilize nuclear sampling, setting the temperature to 0.1 and top\_p to 0.9. We utilize the default RoBERTa-based classifier ToxiGen \cite{liu2019roberta} for identifying toxic generations. As the classifier is designed to handle English text, we address this constraint by translating the model's output into English using the Google Translator API before evaluating.
    
    \item \textbf{\texttt{Bold}}\footnote{\url{https://github.com/amazon-science/bold}} \cite{dhamala2021bold}:  The dataset we utilize for bias analysis. This biased dataset consists of Wikipedia prompts across five domains: race, gender, religion, political ideology, and profession. We also utilize vLLM to enhance efficiency. We exclude prompts that belong to the religious ideology subgroups Hinduism and Atheism due to their limited number of prompts. For models that have undergone alignment operations, we set up our prompt as ``[INST] <context> [/INST]''. We configure the model with a max\_tokens setting of 512 and utilize nuclear sampling, setting the temperature to 0.1 and top\_p to 0.9. We use the Valence Aware Dictionary and Sentiment Reasoner (VADER) \cite{hutto2014vader} to compute the sentiment score for the combined prompt and generation text. Additionally, we translate the model's output into English using the Google Translator API before employing VADER to calculate the sentiment score.
\end{itemize}

\end{document}